\pdfoutput=1
\documentclass[11pt]{article}

\usepackage[preprint]{coling}

\usepackage{times}
\usepackage{latexsym}

\usepackage[T1]{fontenc}
\usepackage[utf8]{inputenc}

\usepackage{microtype}

\usepackage{inconsolata}

\usepackage{graphicx}
\usepackage{amsmath}
\usepackage{booktabs}
\usepackage{multirow}
\usepackage{makecell}
\usepackage{hyperref}
\usepackage{svg}
\usepackage{xurl}
\usepackage{bm}
\usepackage{enumitem}
\usepackage{svg}

\newcommand{\nova}{\textsc{NovAScore}}
\newcommand{\novafull}{\underline{\textbf{No}}velty E\underline{\textbf{v}}aluation in \underline{\textbf{A}}tomicity \underline{\textbf{Score}}}
\newcommand{\acubank}{\textit{ACUBank}}
\newcommand{\novaemoji}{
    \includegraphics[width=15pt]{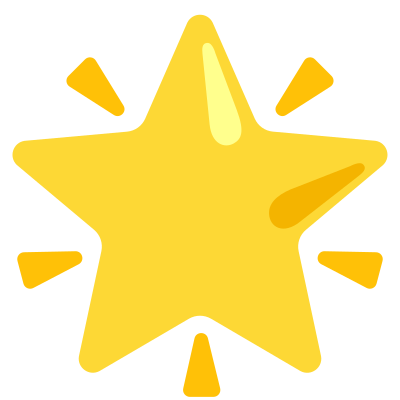}
}

\title{\nova \novaemoji: A New Automated Metric for\\Evaluating Document Level Novelty}

\author{
    \textbf{Lin Ai\textsuperscript{1,2}},
    \textbf{Ziwei Gong\textsuperscript{1,2}},
    \textbf{Harshsaiprasad Deshpande\textsuperscript{1}},
    \textbf{Alexander Johnson\textsuperscript{1}}, \\
    \textbf{Emmy Phung\textsuperscript{1}},
    \textbf{Ahmad Emami\textsuperscript{1}},
    \textbf{Julia Hirschberg\textsuperscript{2}}
    \\
    \textsuperscript{1}Machine Learning Center of Excellence, JPMorgan Chase \& Co.\\
    \textsuperscript{2}Department of Computer Science, Columbia University
    \\
    \{lin.ai, sara.ziweigong, julia\}@cs.columbia.edu, 
}

\begin{document}
\maketitle

\begin{abstract}
The rapid expansion of online content has intensified the issue of information redundancy, underscoring the need for solutions that can identify genuinely new information. Despite this challenge, the research community has seen a decline in focus on novelty detection, particularly with the rise of large language models (LLMs). Additionally, previous approaches have relied heavily on human annotation, which is time-consuming, costly, and particularly challenging when annotators must compare a target document against a vast number of historical documents. In this work, we introduce \textbf{\nova} (\novafull), an automated metric for evaluating document-level novelty. {\nova} aggregates the novelty and salience scores of atomic information, providing high interpretability and a detailed analysis of a document's novelty. With its dynamic weight adjustment scheme, {\nova} offers enhanced flexibility and an additional dimension to assess both the novelty level and the importance of information within a document. Our experiments show that {\nova} strongly correlates with human judgments of novelty, achieving a 0.626 Point-Biserial correlation on the TAP-DLND 1.0 dataset and a 0.920 Pearson correlation on an internal human-annotated dataset.

\end{abstract}
\section{Introduction}
\label{sec:intro}

\begin{figure}[t]
    \centering
    \includegraphics[width=0.4\textwidth]{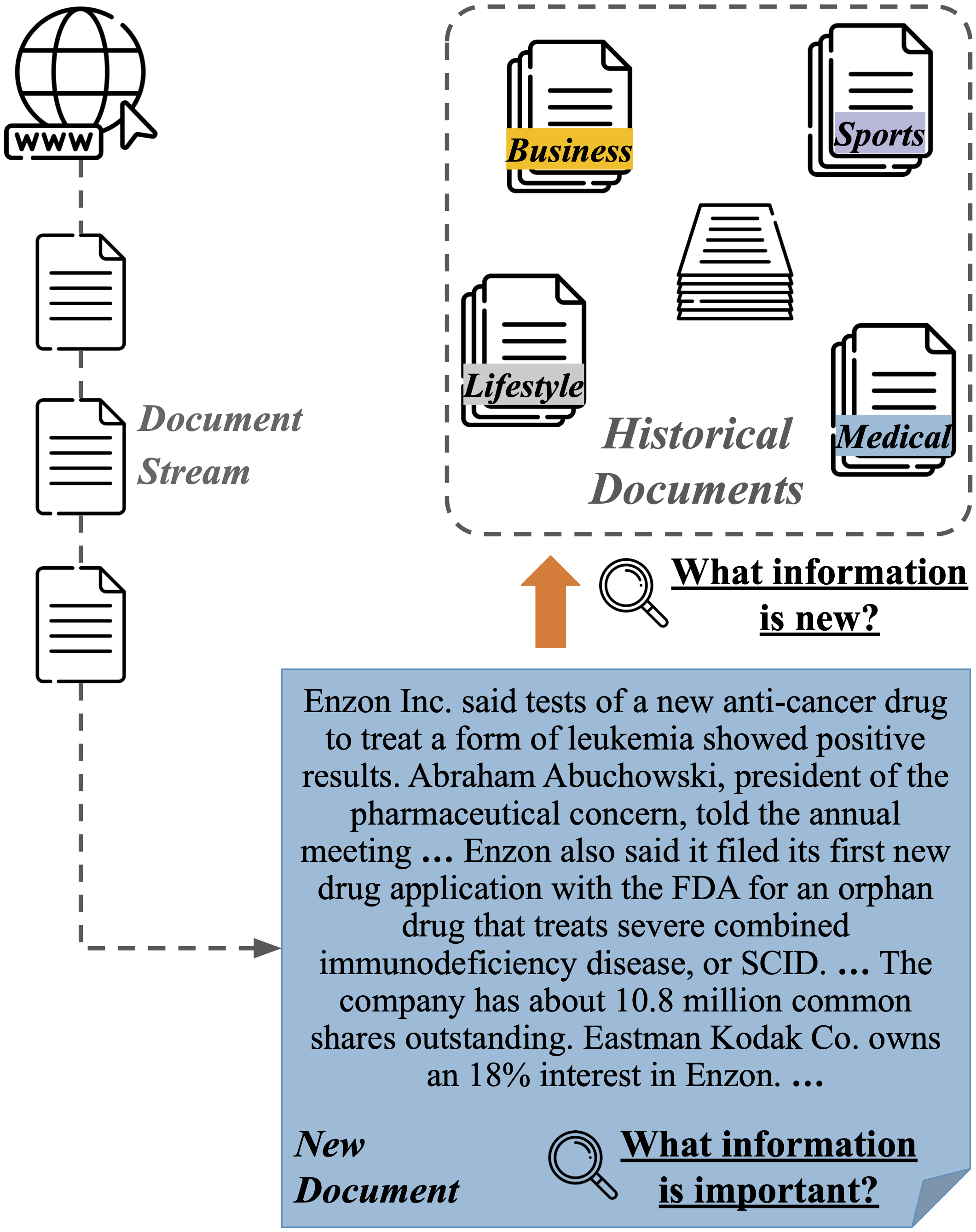}
    \caption{Conceptual illustration of novelty and salient information retrieval in real-world applications.}
    \label{fig:novelty_detection}
    % \vspace{-0.5cm}
\end{figure}

Textual novelty detection has long been a key challenge in information retrieval (IR) \cite{soboroff-harman-2005-novelty}, focusing on identifying text that introduces new, previously unknown information. With the rapid expansion of online content, this issue has become more significant, as redundant information increasingly obstructs the delivery of critical, timely, and high-quality content \cite{ghosal-etal-2022-novelty}. A 2022 Google SEO study revealed that 60\% of internet content is duplicated \cite{seroundtableGoogleSays}. The rise of Large Language Models (LLMs) has further contributed to the generation of artificial and semantically redundant information. Detecting whether a document provides new, relevant, and salient information is crucial for conserving space, saving time, and maintaining reader engagement. 

In addition, recent work \cite{li2024autobencher} introduces novelty as a key metric for benchmark design, noting that performance on existing benchmarks is often highly correlated \cite{liu-etal-2023-question, perlitz-etal-2024-efficient, polo2024tinybenchmarks}. Novelty helps uncover hidden performance patterns and unexpected model behaviors, enabling more dynamic evaluations and the development of higher-quality benchmarks that push the limits of model improvement.

Despite the increasing issue of information redundancy and the growing need for novelty in benchmarking, focus on novelty detection has declined, especially since the rise of LLMs after 2022. Most prior efforts in document-level novelty detection rely on single categorical classification, lacking detailed analysis of what is genuinely new within a document. Additionally, previous work has overlooked the salience of information -- how important each piece is and how it contributes to assessing a document's overall novelty and value. These methods also heavily depend on human annotation, which is time-consuming, costly, and challenging, especially when comparing a target document against many historical documents \cite{ghosal-etal-2018-novelty}, as illustrated in Figure \ref{fig:novelty_detection}. 

\textbf{\textit{Our motivation is twofold:}} \textbf{(a)} to develop a new metric for document-level novelty that offers granular analysis and incorporates the salience of information, and \textbf{(b)} to provide an automated solution that reduces the costs and time associated with manual labeling. Our contributions are as follows:
\begin{itemize}[leftmargin=*,nosep,topsep=0pt]
    % \vspace{-0.2cm}
    \item[1.] We introduce \textbf{\nova}, short for \novafull, an automated metric for evaluating document-level novelty. {\nova} aggregates the novelty and salience scores of atomic content units, providing high interpretability and demonstrating strong correlation with human judgments of novelty.
    % \vspace{-0.2cm}
    \item[2.] We release {\nova} as an open-source tool\footnote{Code will be released soon.}, encouraging further research to expand its applicability and enhance its scalability.
\end{itemize}
\section{Related Work}
\label{sec:related_work}

\paragraph{Novelty Detection}

Textual novelty detection has its roots in early IR research, particularly through the Topic Detection and Tracking (TDT) campaigns. These efforts focused on new event detection by clustering news stories based on similarity thresholds \cite{wayne1997topic, brants2003system}. The task gained further prominence during the Text Retrieval Conferences (TREC) from 2002 to 2004, where sentence-level novelty detection became a focal point \cite{soboroff2003overview, clarke2004overview, soboroff-harman-2005-novelty, schiffman-mckeown-2005-context}. While sentence-level detection was well-researched, it is insufficient for addressing the vast amount of document-level information available on the web today \cite{ghosal-etal-2022-novelty}.

At the document level, \citet{yang2002topic} pioneered the use of topical classification for detecting novelty in online document streams. \citet{zhang2002novelty} introduced redundancy measures to assess document novelty. More recent approaches have explored information entropy measures \cite{dasgupta2016automatic}, deep neural networks \cite{ghosal-etal-2018-novelty}, multi-source textual entailment \cite{ghosal-etal-2022-novelty}, and unsupervised approaches \cite{nair2024predicting} for detecting novelty in documents.

% However, despite the growing issue of information redundancy \cite{ghosal-etal-2022-novelty}, there has been a noticeable decline in focus on novelty detection and evaluation, especially since the emergence of LLMs. Our work aims to fill this gap by leveraging the capabilities of LLMs. Moreover, while previous research has relied entirely on human-annotated data, we propose a fully automated metric for evaluating document-level novelty, which will reduce the time and cost associated with human annotation.

\paragraph{Information Similarity Evaluation} Directly assessing the novelty of information is challenging. However, numerous metrics exist for evaluating \textit{semantic similarity} between pieces of information. A common approach involves using cosine similarity between contextual embeddings, as seen in methods like BertScore \cite{zhangbertscore}, MoverScore \cite{zhao-etal-2019-moverscore}, and BartScore \cite{yuan2021bartscore}. Additionally, Natural Language Inference (NLI) is widely recognized for evaluating information similarity and consistency. It is frequently employed in novelty detection \cite{dagan2022recognizing, ghosal-etal-2022-novelty}, summarization evaluation \cite{liu-etal-2023-towards-interpretable, laban-etal-2022-summac}, and factuality assessment \cite{min-etal-2023-factscore, zha-etal-2023-alignscore, ji2023survey}. Beyond these one-stage metrics, two-stage approaches, such as QA-based methods, are extensively used to evaluate information overlap and faithfulness in both summarization and factuality evaluations \cite{deutsch-etal-2021-towards, zhong-etal-2021-qmsum, goyal2022news, fabbri-etal-2022-qafacteval}. In our work, we utilize and assess all three categories of approaches as a close approximation for identifying semantic-level non-novelty.
\section{\nova}
\label{sec:nova}

\begin{figure*}[ht]
    \centering
    \includegraphics[width=0.95\textwidth]{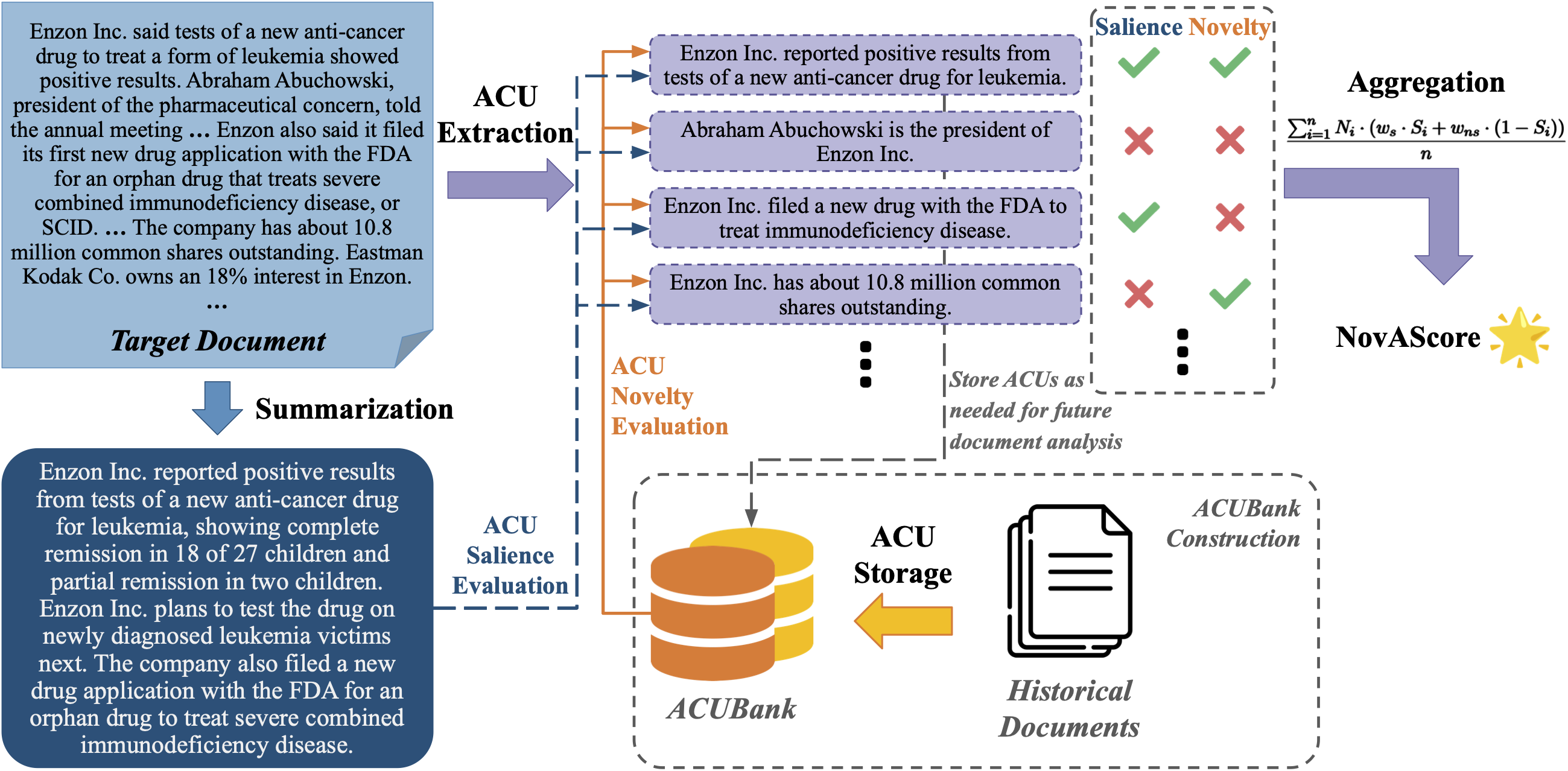}
    \caption{The {\nova} framework. The target document is first decomposed into ACUs. ACU-level novelty is assessed by comparing each ACU against the {\acubank} of historical documents, while salience is determined by whether the ACU is included in the document's summary. The final {\nova} is calculated by aggregating the scores of the ACUs. ACUs can be stored in the {\acubank} for future analysis if necessary.}
    \label{fig:novascore}
    \vspace{-0.2cm}
\end{figure*}

We introduce {\nova}, a new automated method for evaluating the novelty of a target document compared to a series of historical documents. Unlike previous methods that assign a categorical value to the entire document, {\nova} offers an interpretable and granular analysis at the atomicity level. As shown in Figure \ref{fig:novascore}, the process starts by decomposing the target document into \textit{Atomic Content Units} (ACUs). We define an ACU similarly to \citet{min-etal-2023-factscore} (\textit{atomic facts}) and \citet{liu-etal-2023-revisiting}, but with a more holistic perspective -- an elementary information group that combines the minimal number of atomic facts necessary to convey a single message. Each ACU is then evaluated for novelty by comparing it to an {\acubank} of historical documents and assessed for its salience within the document's context. The overall {\nova} is computed by aggregating the novelty and salience scores of all ACUs. After processing, the target document ACUs can be stored in the {\acubank} for future analysis. This approach allows for a high-level assessment of the document’s novelty while precisely identifying new and important information, providing fine-grained interpretability.

\subsection{ACU Extraction and \acubank}
\label{subsec:acu_extraction}
We build on the ideas from \citet{liu-etal-2023-revisiting} and \citet{min-etal-2023-factscore} to extract abstractive ACUs, but unlike their methods, which break down sentences into highly fine-grained units, we extract document-level ACUs directly. This approach better suits our task of novelty detection, which requires a holistic evaluation of new information rather than overly fine details. We frame automatic document-level ACU extraction as a sequence-to-sequence problem \cite{sutskever2014sequence}: $m(D) \rightarrow A$, where $D$ is the input document, $m$ is a language model, and $A$ is the set of generated document-level ACUs.

While previous research has focused on sentence-level novelty \cite{schiffman-mckeown-2005-context, ghosal-etal-2022-novelty}, we choose ACU-level analysis to better handle complex, information-dense sentences and to maintain context by considering messages that may span multiple sentences.

To efficiently evaluate the novelty of target ACUs, we construct an \textbf{{\acubank}} -- a collection of databases that store ACUs from historical documents -- allowing for quick similarity searches at minimal computational cost without the need for real-time relevant content retrieval. The databases are built by indexing ACUs using SentenceBERT embeddings \cite{reimers-gurevych-2019-sentence}. For each ACU, the most relevant historical ACUs are rapidly retrieved via semantic cosine similarity to assess novelty. To speed up searches, the {\acubank} is organized into multiple databases, each containing ACUs from specific \textit{clusters}, so a target document is only searched within its cluster, significantly narrowing the search scope.

\subsection{ACU Novelty Evaluation}
\label{subsec:acu_novelty}
We approximate non-novel information assessment using three common information similarity evaluators: embedding cosine similarity, NLI, and QA, as discussed in Section \ref{sec:related_work}. These evaluators assess ACU novelty, treating it as a binary task -- determining whether the information is new or not.

\paragraph{Cosine similarity} provides a straightforward approach to evaluating ACU novelty. We compare each target ACU with historical ACUs from the {\acubank}. If any historical ACU exceeds a set similarity threshold with the target ACU, it is classified as non-novel, indicating likely repetition; and vice versa. This method efficiently assesses overlap, making it a practical tool for novelty detection.

\paragraph{NLI} is based on the principle that a premise $P$ entails a hypothesis $H$ if a typical human reader would conclude that $H$ is most likely true after reading $P$ \cite{dagan2005pascal}. In the context of novelty detection, this means that \textit{if one or more entailing premises are found for a given hypothesis, the content of that hypothesis is considered not new} \cite{bentivogli2011seventh}. To evaluate ACU-level novelty, we concatenate the most relevant historical ACUs into a single premise and compare it against the target ACU as the hypothesis. If the historical content entails the target ACU, it is classified as non-novel; otherwise, it is considered novel.

\paragraph{QA-based} approach is widely used to evaluate information overlap by representing reference information as question-answer pairs, with recall assessed by how well the candidate text answers these questions \cite{deutsch-etal-2021-towards, zhong-etal-2021-qmsum}. We adapt this method in reverse: for each target ACU, we generate questions where the target ACU itself is the answer. If any historical ACUs can answer these questions, the target ACU is considered non-novel; otherwise, it is novel. We generate three questions per ACU, focusing on \textit{named entities} and \textit{noun phrases} \cite{deutsch-etal-2021-towards}. The answers derived from historical ACUs are consolidated into a single sentence and compared to the target ACU. If the consolidated answer has a cosine similarity of 0.85 or higher with the target ACU, it is classified as non-novel. The rationale for this threshold is detailed in Appendix \ref{subsubsec:similarity_threshold}.

\subsection{ACU Salience Evaluation}
\label{subsec:acu_salience}
Not all information in a document is equally important. For instance, as shown in Figure \ref{fig:novascore}, the primary focus of the target document is Enzon Inc.‘s positive results for a new medication, while the company’s ownership structure, briefly mentioned later, is less significant within the pharmaceutical domain. Therefore, when evaluating novelty, it is essential to prioritize the most important content to ensure an accurate assessment.

To determine the salience of each ACU, we compare it to the document's summary. The underlying assumption is that a high-quality summary should include all and only the essential information from the document. Therefore, we formulate ACU salience evaluation as a binary classification problem: whether or not an ACU is mentioned in the document's summary.

\subsection{ACU Scores Aggregation}
\label{subsec:acu_aggregation}
When aggregating ACU scores to compute the overall {\nova} of a document, it is essential to assign higher weights to salient ACUs to accurately reflect their importance. To achieve this, we implement a \textbf{dynamic weight adjustment} scheme based on the following principles:

\begin{itemize}[leftmargin=*,nosep,topsep=0pt]
    % \vspace{-0.2cm}
    \item[1.] \textbf{Salience Emphasis at Low Salience Ratio:} When the ratio of salient ACUs is low, each salient ACU is assigned a significantly higher weight compared to non-salient ACUs. This ensures that the final score is not overly influenced by the novelty of less important content.
    % \vspace{-0.2cm}
    \item[2.] \textbf{Non-Salience Boost at High Salience Ratio:} When the proportion of salient ACUs is high, the weights of non-salient ACUs are increased slightly to ensure they still contribute meaningfully to the overall score.
    % \vspace{-0.2cm}
    \item[3.] \textbf{Consistent Prioritization:} Salient ACUs consistently receive higher weights than non-salient ACUs, regardless of their proportion.
    % \vspace{-0.2cm}
\end{itemize}

\begin{figure}[hb]
    \centering
    \includegraphics[width=0.48\textwidth]{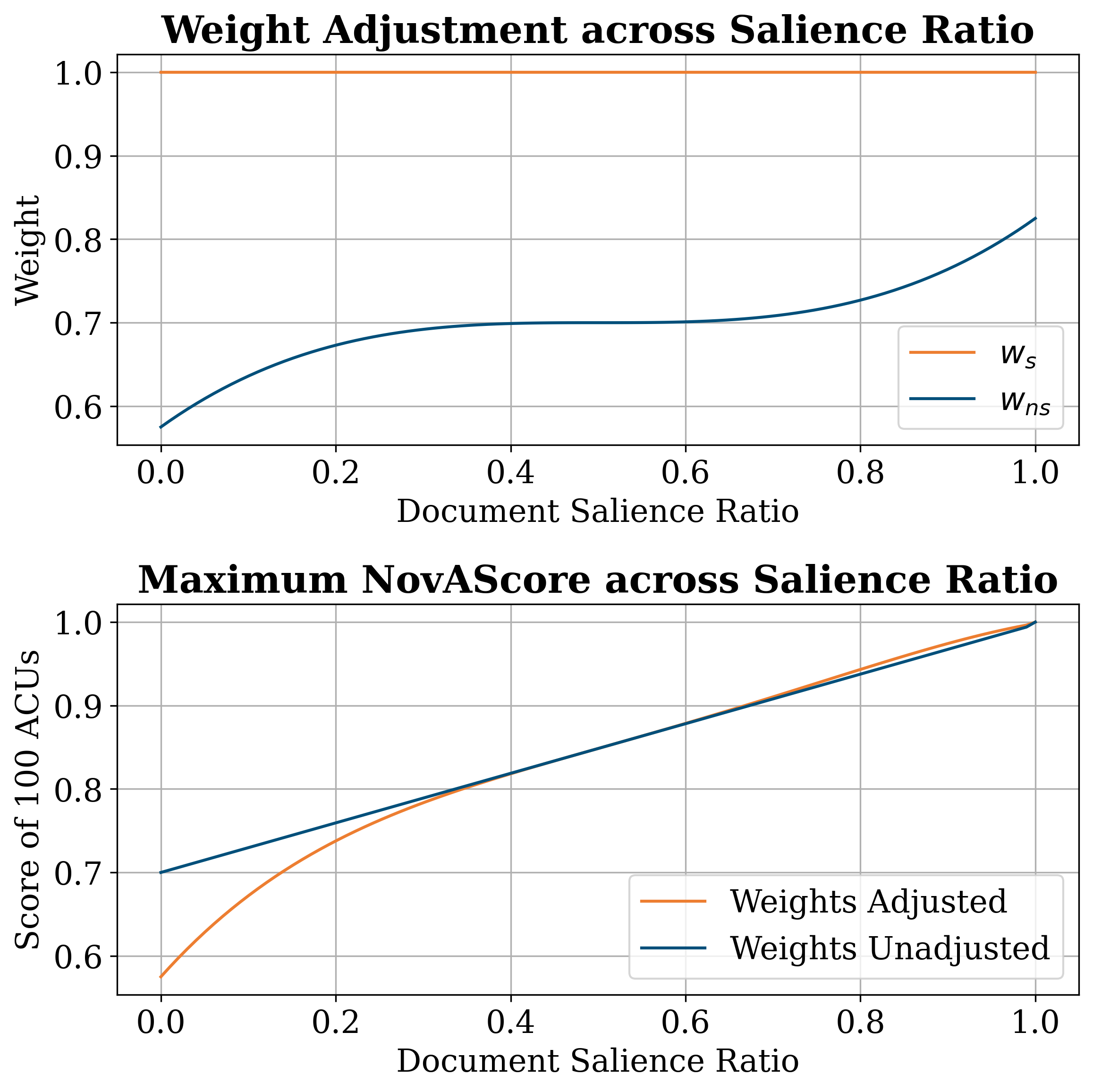}
    \caption{The top plot shows the weights for salient ($w_s$) and non-salient ($w_{ns}$) ACUs across different salience ratios with dynamic weight adjustment. The bottom plot compares the maximum {\nova} of 100 ACUs, with and without weight adjustment. Both plots utilize $\alpha=1$, $\beta=0.5$, and $\gamma=0.7$.}
    \vspace{-0.2cm}
    \label{fig:weights}
\end{figure}

To implement these principles, we set the weight for salient ACUs as $w_s = 1$ and dynamically adjust the weight of non-salient ACUs using a cubic function: $w_{ns} = \text{\textit{min}}(w_s, \alpha (p_s - \beta)^3 + \gamma)$, where $p_s$ is the salience ratio of the document. The hyperparameters $\alpha$, $\beta$, and $\gamma$ shape the cubic curve, with $\alpha$ controlling steepness and $\beta$ and $\gamma$ adjusting the midpoints on the $x$ and $y$ axes. These hyperparameters determine the devaluation of non-salient ACUs and the adjustment for extreme salience ratios, which can vary depending on the datasets and applications. Further details are in Appendix \ref{subsubsec:weight_appendix}. Figure \ref{fig:weights} shows the impact of the weight adjustment scheme: as the salience ratio shifts to very low or high, non-salient ACU weights adjust more rapidly. This adjustment ensures that documents with low salience ratios have a lower maximum {\nova}, giving less value to documents with less salient information. 

% The advantage of this weight adjustment scheme lies in its flexibility to control and incorporate both important and less important information when evaluating the overall novelty of a document. This provides {\nova} with an additional dimension, allowing it to assess not only the level of novelty but also the worthiness of the information within a target document. The detailed explanation and rationale behind these hyperparameter choices is detailed in Appendix \ref{subsubsec:weight_appendix}.

The {\nova} of a document is then:

\begin{small}
\vspace{-0.2cm}
\begin{align*}
    \nova = \frac{\sum_{i=1}^n N_i \cdot (w_s \cdot S_i + w_{ns} \cdot (1 - S_i))}{n}
\end{align*}
\end{small}

where $N_i$ and $S_i$ represent the binary novelty and salience of the $i$-th ACU, respectively, and $n$ denotes the total number of ACUs.
\section{Experiments}
\label{sec:experiments}
In this section, we evaluate the effectiveness of {\nova} by examining its correlation with human judgments of novelty.

\subsection{How Well Does {\nova} Align with Human Judgments of Novelty?}
\label{subsec:corr}
We begin by analyzing how closely {\nova} aligns with human judgments of novelty on a broad scale, specifically by examining its correlation with human-annotated document-level novelty.

\begin{table}[hb]
\centering
\small
    \begin{tabular}{lcc}
    \toprule
    \# & \textbf{TAP-DLND 1.0} & \textbf{APWSJ} \\
    \cmidrule{2-3}
    Novel & 250 & 259 \\
    Non-Novel & 250 & 241 \\
    \textit{Total} & \textit{500} & \textit{500} \\
    \bottomrule
    \end{tabular}
\caption{Statistics of dataset used for experiments.}
\vspace{-0.2cm}
\label{tab:data_stats}
\end{table}

\paragraph{Datasets} 
We utilize the following two datasets, which, to the best of our knowledge, are among the few publicly available in the news domain:

\begin{itemize}[leftmargin=*,nosep,topsep=0pt]
    % \vspace{-0.2cm}
    \item[1.] \textbf{TAP-DLND 1.0} \cite{ghosal-etal-2018-tap}: This dataset contains 2,736 human-annotated novel documents and 2,704 non-novel documents, all clustered into specific categories. Each novel or non-novel document is annotated against three source documents.
    % \vspace{-0.2cm}
    \item[2.] \textbf{APWSJ} \cite{zhang2002novelty}: This dataset comprises 10,833 news articles from the Associated Press (AP) and Wall Street Journal (WSJ) corpora, covering 33 topics. The documents are chronologically ordered and annotated into three categories: \textit{absolutely redundant}, \textit{somewhat redundant}, and \textit{novel}. Of these, 7,547 are novel, 2,267 are somewhat redundant, and 1,019 are absolutely redundant.
    % \vspace{-0.3cm}
\end{itemize}

For both datasets, we sample 500 documents. In TAP-DLND 1.0, we randomly select 500 documents across clusters. For APWSJ, where documents are chronologically sorted and annotated for novelty relative to earlier ones, we sequentially select a balanced set of 500 non-novel and novel documents. Table \ref{tab:data_stats} provides the dataset statistics used in our experiments.

\begin{table*}[ht]
\centering
\resizebox{0.95\textwidth}{!}
{
    \begin{tabular}{lccccccc}
    \toprule
    \textbf{Dataset $\rightarrow$} & \multicolumn{3}{c}{\textbf{TAP-DLND 1.0}} && \multicolumn{3}{c}{\textbf{APWSJ}} \\
    \textit{{Novelty Evaluator}} $\rightarrow$ & CosSim & NLI & QA && CosSim & NLI & QA \\
    \cmidrule{2-4} \cmidrule{6-8}
    \textbf{Correlation $\downarrow$} &  &  &  &&  &  &  \\
    Point-Biserial & \underline{0.545}$_{(4.9e-40)}$ & \textbf{0.626}$_{(9.2e-56)}$ & 0.508$_{(2.3e-33)}$ && \underline{0.447}$_{(6.5e-26)}$ & \textbf{0.476}$_{(1.2e-29)}$ & 0.422$_{(2.2e-22)}$ \\
    Spearman & \underline{0.555}$_{(1.1e-41)}$ & \textbf{0.622}$_{(9.2e-55)}$ & 0.497$_{(8.1e-32)}$ && \underline{0.446}$_{(9.0-26)}$ & \textbf{0.482}$_{(1.8e-30)}$ & 0.439$_{(2.9e-24)}$ \\
    Kendall & \underline{0.460}$_{(2.8e-35)}$ & \textbf{0.510}$_{(8.0e-44)}$ & 0.409$_{(5.4e-28)}$ && \underline{0.358}$_{(2.6e-24)}$ & \textbf{0.395}$_{(4.8e-27)}$ & 0.353$_{(5.2e-23)}$ \\
    \bottomrule
    \end{tabular}
}
\caption{The correlations (statistics$_{(p\text{-value})}$) between {\nova} and human annotations on the TAP-DLND 1.0 and APWSJ datasets, using different novelty evaluators and correlation metrics.}
\vspace{-0.2cm}
\label{tab:corr}
\end{table*}

\paragraph{Setup and Implementation}
We utilize GPT-4o across all modules, including ACU extraction, document summarization, salient ACU selection, and both NLI and QA-based novelty evaluation. Details on the prompts are provided in Appendix \ref{subsec:prompts}.

The {\acubank} is implemented using FAISS\footnote{\href{https://ai.meta.com/tools/faiss/}{Link to FAISS.}} for fast similarity search and efficient indexing. Each ACU is indexed by its sentence embedding from the pre-trained SentenceBERT\footnote{\href{https://huggingface.co/sentence-transformers/all-mpnet-base-v2}{Model card of all-mpnet-base-v2}.}. In TAP-DLND 1.0, we create separate databases for each of the 223 clusters. For APWSJ, documents are processed chronologically into a unified database.

To retrieve relevant historical ACUs from the {\acubank}, we select the top-5 ACUs with a cosine similarity of 0.6 or higher. The rationale for this threshold is detailed in Appendix \ref{subsubsec:similarity_threshold}. If any meet this threshold, the ACU is considered non-novel when using the cosine similarity novelty evaluator. For NLI or QA novelty evaluators, these similar ACUs are concatenated to form the premise for NLI or the context for QA, further assessing the ACU's novelty. 

We use different hyperparameters for each dataset in the dynamic weight adjustment to account for ACU salience when calculating the overall {\nova}. These parameters control the devaluation of non-salient ACUs and adjust for extreme salience ratios, varying by dataset. For TAP-DLND 1.0, we use $\alpha=0$, $\beta=0.5$, and $\gamma=1$ (no adjustment). For APWSJ, we use $\alpha=1$, $\beta=0.5$, and $\gamma=0.7$. The rationale for these choices is detailed in Appendix \ref{subsubsec:weight_appendix}.

\paragraph{Metrics}
We employ \textbf{Point-Biserial}, \textbf{Spearman}, and \textbf{Kendall} to evaluate the relationship between the {\nova} and human annotations of document-level novelty. Point-Biserial is a special case of Pearson correlation that compares continuous variables with binary variables. For the TAP-DLND 1.0 dataset, where human annotations are binary, we assign a label of 1 to \textit{novel} and 0 to \textit{non-novel} for calculating correlations. In contrast, the APWSJ dataset contains three classes: we assign \textit{novel} a value of 1, \textit{somewhat redundant} a value of 0.5, and \textit{absolute redundant} a value of 0 for the Spearman and Kendall correlations. For Point-Biserial correlation on APWSJ, we set both \textit{absolute redundant} and \textit{somewhat redundant} to 0.

\paragraph{Results}
As shown in Table \ref{tab:corr}, across both datasets and using all novelty evaluators, {\nova} demonstrates a moderate to strong correlation with human judgments of document-level novelty, as indicated by the correlation cutoffs detailed in Appendix \ref{subsec:corr_interpretation}. These correlations are statistically significant, with $p$-values ranging from $10^{-22}$ to $10^{-56}$, indicating the robustness of {\nova} in aligning with human perceptions of novelty.

Among the different evaluators, the NLI novelty evaluator consistently outperforms the others, showing a particularly strong correlation with human annotations. Notably, on the TAP-DLND 1.0 dataset, the {\nova} with the NLI novelty evaluator achieves a Point-Biserial of 0.626, a Spearman of 0.622, and a Kendall of 0.510, all signifying a strong alignment with human judgment.
\subsection{Can {\nova} Capture Granular Insights at the ACU Level?}
\label{subsec:human}

In addition to broad document-level novelty analysis, we also assess {\nova}'s reliability on a granular scale by examining its alignment with human judgments of novelty at the ACU level.

\paragraph{Human Annotation}
Since existing public datasets only provide single categorical labels at the document level without fine-grained annotations, we curate and annotate a new dataset for this purpose. We manually select 32 news articles, clustered into 8 topics. Within each topic, the documents are sorted in chronological order, and we extract ACUs using GPT-4o, following the strategy described in Sections \ref{subsec:acu_extraction}. Human annotators evaluate each ACU based on four labels: correctness (logical and factual consistency), redundancy (non-informativeness or intra-document non-novelty), novelty, and salience. The full annotation instructions and label schema are provided in Appendix \ref{subsec:annotation_instruction}. Two annotators independently perform the entire annotation task. After completing the annotations, the annotators meet to discuss and resolve any conflicting annotations, ensuring consensus on the final labels. Further discussion on annotation quality is presented in Appendix \ref{subsec:annotation_quality}.

\paragraph{Novelty Evaluator Performance}
We compare the performance of each novelty evaluator against human annotations of ACU-level novelty. As shown in Table \ref{tab:novelty_evaluator}, all novelty evaluators achieve strong classification results, with the NLI-based evaluator leading with an accuracy of 0.94.

\begin{table}[ht]
\centering
\resizebox{0.42\textwidth}{!}
{
    \begin{tabular}{lccc}
    \toprule
    \textbf{Novelty Evaluator} $\rightarrow$ & CosSim & NLI & QA \\
    \cmidrule{2-4}
    \textbf{Metric} $\downarrow$ \\
     Accuracy & 0.83 & \textbf{0.94} & \underline{0.91} \\
    Macro F1 & 0.71 & \textbf{0.84} & \underline{0.80} \\
    \bottomrule
    \end{tabular}
}
\caption{Novelty evaluator performance.}
\vspace{-0.2cm}
\label{tab:novelty_evaluator}
\end{table}

\paragraph{{\nova} vs Human Judgments}
We aggregate ACU-level scores to compute the document-level novelty score, resulting in the following {\nova} variants: \textbf{(1)} {\nova}$_{\text{human}}$, using human-annotated novelty and salience labels, and \textbf{(2)} {\nova}$_{\text{CosSim}}$, {\nova}$_{\text{NLI}}$, and {\nova}$_{\text{QA}}$, which are fully automated versions utilizing their respective novelty evaluators and GPT salience evaluator. For all variants, we apply weight adjustment parameters of $\alpha=1$, $\beta=0.5$, and $\gamma=0.7$, as used for the APWSJ dataset. We also compute all variants without incorporating salience to better assess the performance of the novelty evaluators.

We then use Pearson correlation to evaluate the relationship between {\nova}$_{\text{human}}$ and the three automated variants, as all produce continuous scores with nearly linear distributions. As shown in Table \ref{tab:corr_human}, all automated variants demonstrate strong to very strong correlations, with {\nova}$_{\text{NLI}}$ achieving the highest Pearson correlation of 0.920 without salience and 0.835 with salience. The discrepancy between human-annotated and GPT-selected salient ACUs, as discussed in Section \ref{subsec:gpt_performance}, slightly degrades the correlation statistics when salience is incorporated, which is expected. Despite this, the fully automated {\nova} with salience included still shows a very strong correlation with human judgments, indicating that GPT-4o is a reasonable estimator of information salience. The full correlation results between the three automated variants and {\nova}$_{\text{human}}$ are detailed in Table \ref{tab:corr_human_full} in Appendix \ref{subseec:corr_human_full}.

\begin{table}[ht]
\centering
\resizebox{0.48\textwidth}{!}
{
    \begin{tabular}{lccc}
    \toprule
     & \small{{\textbf{\nova}}$_{\text{\textbf{CosSim}}}$} & \small{{\textbf{\nova}}$_{\text{\textbf{NLI}}}$} & \small{{\textbf{\nova}}$_{\text{\textbf{QA}}}$} \\
    \cmidrule{2-4}
    \textbf{Salience} $\downarrow$ \\
    w/o & 0.748$_{(8.5e-07)}$ & \textbf{0.920}$_{(9.6e-14)}$ & \underline{0.843}$_{(2.9e-09)}$ \\
    w/ & 0.722$_{(3.1e-06)}$ & \textbf{0.835}$_{(2.9e-09)}$ & \underline{0.779}$_{(2.4e-07)}$ \\
    \bottomrule
    \end{tabular}
}
\caption{The Pearson correlations (statistics$_{(p\text{-value})}$) between {\nova}$_{\text{human}}$ and {\nova} with different novelty evaluators.}
\vspace{-0.2cm}
\label{tab:corr_human}
\end{table}

These results underscore the effectiveness of {\nova} in capturing document-level novelty while also providing fine-grained interpretability at the atomic level, making it a reliable tool for assessing document-level novelty.
\subsection{How Does Dynamic Weight Adjustment Enhance Novelty Evaluation?}
\label{subsec:weight}

\begin{table}[hb]
\centering
\resizebox{0.48\textwidth}{!}
{
    \begin{tabular}{lcc}
    \toprule
    & {\textbf{\nova}}$_{\text{\textbf{NLI}}}$ & {\textbf{\nova}}$_{\text{\textbf{NLI}}}$ \textbf{w/o WA} \\
    \cmidrule{2-3}
    \textbf{Correlation $\downarrow$} \\
    Point-Biserial & \textbf{0.476}$_{(1.2e-29)}$ & 0.442$_{(2.2e-25)}$ \\
    Spearman & \textbf{0.482}$_{(1.8e-30)}$ & 0.468$_{(1.4e-28)}$ \\
    Kendall & \textbf{0.395}$_{(4.8e-27)}$ & 0.393$_{(1.4e-25)}$ \\
    \bottomrule
    \end{tabular}
}
\caption{The correlations (statistics$_{(p\text{-value})}$) between {\nova}$_{\text{NLI}}$ and human annotations on APWSJ with and without weight adjustment (WA).}
\vspace{-0.2cm}
\label{tab:weight}
\end{table}

As discussed in Section \ref{subsec:corr} and Appendix \ref{subsubsec:weight_appendix}, the dynamic weight adjustment scheme is designed to ensure that the overall {\nova} score reflects both the novelty and importance of the information. The magnitude and rate of adjustment, controlled by the hyperparameters, vary depending on the specific dataset and its standards. In datasets like APWSJ, where document-level novelty is determined with nuanced considerations of redundancy and individual perception differences, the concept of information salience is implicitly included in the final novelty label. As shown in Table \ref{tab:weight}, incorporating weight adjustment on APWSJ consistently results in higher correlation values across all metrics. Conversely, on datasets like TAP-DLND 1.0, where novelty labels are strict binary cutoffs reflecting only whether there is sufficient new information, adding weight adjustment does not necessarily improve the correlation.

The strength of this weight adjustment scheme lies in its flexibility to emphasize both important and less critical information when evaluating a document's overall novelty, tailored to the needs of the specific application. This provides {\nova} with an additional dimension, enabling it to assess not only the level of novelty but also the worthiness of the information within a target document.

\section{Discussion}
\label{sec:discussion}

\subsection{GPT-4o Performance and Reliability}
\label{subsec:gpt_performance}

\paragraph{ACU Extraction}
As introduced in Section \ref{subsec:human}, we collect correctness and redundancy labeled on GPT-generated ACUs during human annotation. Results reveal that none of the GPT-4o generated ACUs are labeled as incorrect by the annotators, and only 0.1\% are considered redundant. These findings indicate that GPT-4o is highly reliable in generating high-quality ACUs.

Currently, no public datasets are designed for abstractive document-level ACU extraction. The closest are summarization datasets, which focus on key information and miss non-salient ACUs. Thus, our evaluation prioritizes precision over recall, leaving full ACU extraction and novelty recall for future work (see Limitations section).

\paragraph{Salience Evaluation}
Recent studies suggest that LLM-generated summaries are often on par with human-written ones \cite{zhang-etal-2024-benchmarking}, supporting our confidence in GPT-4o's ability to evaluate salience. We compare GPT-selected salient ACUs with human-annotated salience labels as outlined in Section \ref{subsec:human}, and find that GPT achieves a macro F1 score of 0.6. This discrepancy may result from the different conditions: human annotators determine salience in real-time without summaries, making the task more challenging, while GPT-4o can reference the generated summary. Additionally, salience is inherently subjective, making it difficult to standardize. Despite these factors, GPT-4o's performance in salience evaluation is satisfactory.

\subsection{Cost}
\label{subsec:cost}

\begin{table}[ht]
\centering
\resizebox{0.48\textwidth}{!}
{
    \begin{tabular}{lcc}
    \toprule
    \textbf{Dataset $\rightarrow$} & \textbf{APWSJ} & \textbf{TAP-DLND 1.0} \\
    \cmidrule{2-3}
    \textbf{Module $\downarrow$} \\
    ACU Extraction \& Salience & 1.7M & 1.5M \\
    NLI Novelty Evaluator & 0.8M & 0.9M \\
    QA Novelty Evaluator & 2.0M & 1.7M \\
    \bottomrule
    \end{tabular}
}
\caption{Tokens Utilized in GPT-4o Calls for Each Module. “ACU Extraction \& Salience” refers to the one-pass call that performs both ACU extraction and salience evaluation.}
\vspace{-0.2cm}
\label{tab:cost}
\end{table}

\noindent We report the token usage of {\nova} during GPT-4o calls, as shown in Table \ref{tab:cost}. ``ACU Extraction \& Salience'' refers to the one-pass call that handles both tasks, with the detailed prompt in Section \ref{subsec:prompts}. The embedding cosine similarity evaluator doesn't require GPT calls, making it cost-effective, especially for large-scale evaluations.

The QA novelty evaluator consumes about twice as many tokens as the NLI evaluator due to its two-step process: question generation followed by question answering. In addition, although QA-based methods are effective in other tasks like summarization evaluation (as discussed in Sections \ref{sec:related_work} and \ref{subsec:acu_novelty}), they don't perform as well as NLI and sometimes even embedding cosine similarity on novelty detection. Therefore, \textbf{if budget is not a concern, we recommend using the NLI novelty evaluator for its strong performance}. Alternatively, embedding cosine similarity offers a good balance between cost and effectiveness.

\subsection{Scalability}
\label{subsec:scalability}

\begin{figure}[b]
    \centering
    \includegraphics[width=0.48\textwidth]{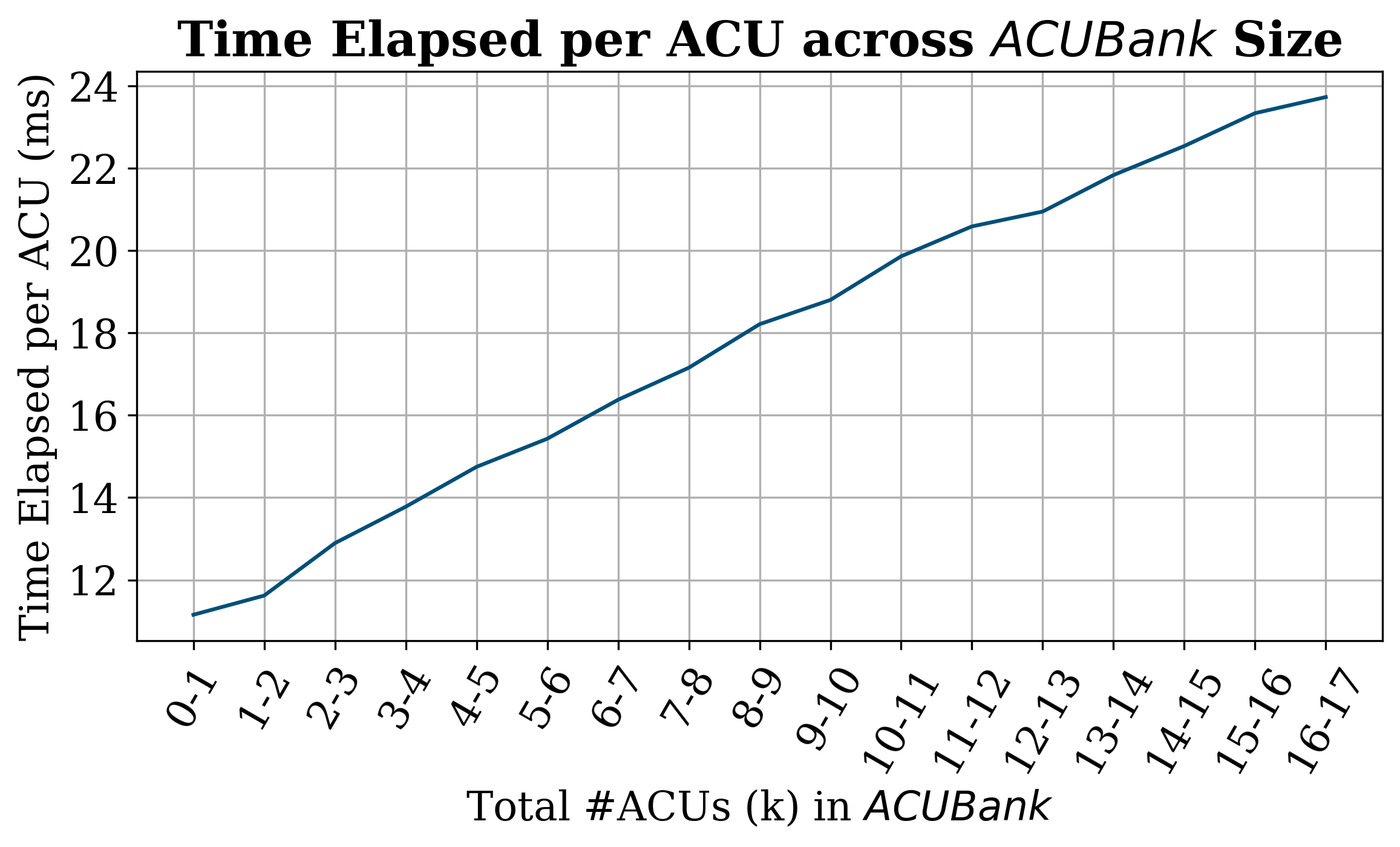}
    \caption{Search time for similar ACUs per ACU at varying {\acubank} sizes with a single database.}
    \vspace{-0.2cm}
    \label{fig:search_time}
\end{figure}

We examine the time required to search for similar ACUs across different {\acubank} sizes. As shown in Figure \ref{fig:search_time}, search time increases linearly with the size of the {\acubank}. To improve scalability, \textbf{clustering documents or ACUs and creating separate databases within the {\acubank} for each cluster would reduce search space and time}.

We do not report the average latency of GPT API calls, as various factors -- such as usage time and network conditions -- can affect this. However, we acknowledge that potential API lags could increase the framework's runtime. Replacing some modules with locally hosted smaller models, like fine-tuned open-source NLI models, could mitigate these delays and enhance efficiency.

\subsection{Applications}
\label{subsec:applications}
Novelty detection in NLP has broad applications across various tasks, including plagiarism detection \cite{gipp2014citation}, news event tracking \cite{ghosal-etal-2018-tap}, scientific novelty detection \cite{gupta2024scind, kelty2023don}, and misinformation detection \cite{qin2016spotting, Ai_Chen_Gong_Guo_Hooshmand_Yang_Hirschberg_2021}. Furthermore, recent work \cite{li2024autobencher} introduces novelty as a key metric for benchmark design, revealing hidden performance patterns and unexpected model behaviors, which enhances evaluations and drives the creation of higher-quality benchmarks that advance model development. Despite its wide-ranging utility, this area has not received sufficient attention. Our work aims to address this gap and push forward the research on novelty detection.
\section{Conclusions and Future Work}

In this work, we introduce \textbf{\nova}, an automated metric for evaluating document-level novelty. {\nova} considers novelty and salience at the atomic level, providing high interpretability and detailed analysis. By incorporating information salience and a dynamic weight adjustment scheme, {\nova} offers enhanced flexibility and an additional dimension, allowing it to assess not only the level of novelty but also the worthiness of the information when evaluating the overall novelty of a document. Our experiments on both public datasets and an internal human-annotated dataset demonstrate that {\nova} strongly correlates with human judgments of novelty, validating its effectiveness and reliability.

% Looking ahead, we aim to explore ways to improve the cost and scalability of {\nova} by integrating open-source LLMs and smaller models to replace GPT-4o in each module. This would reduce dependency on proprietary systems and enhance the accessibility of our framework. 
% Additionally, we plan to construct a publicly available human-annotated dataset to further support research in document-level novelty detection.

Looking ahead, we aim to explore ways to improve the cost and scalability of {\nova} by integrating open-source LLMs and smaller models to replace GPT-4o in each module. This would reduce dependency on proprietary systems and enhance the accessibility of our framework. 

Additionally, as outlined in Section \ref{subsec:applications}, {\nova} serves as a foundation for various tasks and applications. We encourage further research to expand its use across more fields, and believe its potential in novelty detection and model evaluation will have a strong impact on the research community.

\newpage
\section*{Limitations}

\paragraph{Internal Human Annotated Data Restriction} 
One limitation of our study is that the human-annotated data discussed in Section \ref{subsec:human} is internal and proprietary, which means we cannot provide additional information about the specific content or characteristics of this data, nor can we release it for public use. However, we do provide complete annotation instructions and schema in Appendix \ref{subsec:annotation_instruction}. Looking ahead, we plan to construct a publicly available human-annotated dataset to address this limitation and support future research in this area.

\paragraph{ACU Extraction Recall Evaluation}
Another limitation of our current approach is the challenge of evaluating the completeness of extracted ACUs in terms of covering the entire content of articles. Currently, there are no public datasets specifically designed for abstractive document-level ACU extraction. The most relevant datasets available are those used for summarization, where documents are paired with human-written summaries. However, these datasets are not ideal for evaluating non-salient ACUs, as summaries typically focus only on the most important information. Similar to \citet{min-etal-2023-factscore}, we rely on machine-generated atomic information as part of our pipeline. Consequently, our evaluation emphasizes precision rather than recall. We acknowledge this limitation and plan to address it in future work, where we aim to conduct a more comprehensive assessment of ACU extraction and novelty recall.

\bibliography{anthology, custom}

\appendix

\newpage
\section{Experiment Details}
\label{sec:appendix}

\subsection{GPT Prompt Templates}
\label{subsec:prompts}
We provide the detailed prompt templates we use in the GPT calls in this section.

\subsubsection{ACU Extraction and Salient ACU Selection Prompt}
\label{subsubsec:acu_extraction_prompt}
\noindent \textbf{INSTRUCTION:}\\
1. First, extract the list of all atomic content units (ACUs) from a given document. An ACU is an elementary information unit in the document that does not require further division. When identifying any named entity, temporal entity, location entity, or attribute, avoid using indirect references. Instead, specify the actual entity, attribute, or noun directly. For example, replace `this company' with the actual name of the company, `this location' with the actual location name, `it' with the actual subject being referred, etc. \\
2. Then, summarize the given document. \\
3. Finally, using the summary, identify the most salient ACUs from the full list of ACUs. The salient ACUs should be those explicitly mentioned in the summary. \\
Output the response in JSON format:\\
\{"all\_acus": "array of ACU strings", "summary": "document summary", "salient\_acus": "array of salient ACU strings"\}
\newline\newline
\textbf{Example 1:}\\
\#\#\#Document: \{example document\}\\
\#\#\#Output: \{example output\}
\newline\newline
\#\#\#Document: \{input document\}\\
\#\#\#Output: 

\subsubsection{NLI Novelty Evaluator Prompt}
\label{subsubsec:nli_prompt}
\textbf{INSTRUCTION:} For each given premise-hypothesis pair, perform Natural Language Inference (NLI) to determine whether the hypothesis should be classified as `entailment', `contradiction', or `neutral' based on the information provided in the premise.\\
Output the response in JSON format:\\
\{"nli\_results": "array of NLI results in the following format: [\{\{"id": int, "nli": "entailment"|"contradiction"|"neutral"\}\}]"\}\\

\noindent ==============\\
\noindent \textbf{EXAMPLE:}\\
\#\#\#Premise 1: ABC Bank reported a significant drop in profits for the second quarter due to rising loan defaults. The bank's CEO mentioned the challenging economic environment as a key factor.\\
\#\#\#Hypothesis 1: ABC Bank's profits declined in the second quarter because of increased loadn defaults.

\noindent \#\#\#Premise 2: Global oil prices surged by 5\% on Monday following geopolitical tensions in the Middle East. Analysts predict that the prices may continue to rise if the situation escalates.\\
\#\#\#Hypothesis 2: Oil price decreased despite tensions in the Middle East.

\noindent \#\#\#Premise 3: The ECB decided to maintain its current monetary policy stance, keeping interest rates unchanged.\\
\#\#\#Hypothesis 3: The ECB's decision will impact the foreign exchange rates of the Euro.\\

\noindent \#\#\#Output:\\
\{"nli\_results": [\{"id": 1, "nli": "entailment"\}, \{"id": 2, "nli": "contradiction"\}, \{"id": 3, "nli": "neutral"\}]\}
\newline\newline
\noindent ==============\\
\{premise (similar ACUs) hypothesis (target ACU) pairs\}\\
\noindent \#\#\#Output:

\subsubsection{QA Novelty Evaluator Prompt}
\label{subsubsec:qa_prompt}
\textbf{\underline{Question Generation Prompt}}\\

\noindent \textbf{INSTRUCTION:} For each given sentence, generate three distinct questions that correspond to the named-entities and noun phrases found in this sentence, and use the sentence as the answer.\\
Output the response in JSON format:\\
\{"questions\_list": "list of question arrays in the format: [[question\_str, ...], [question\_str, ...], ...]"\}\\

\noindent ==============\\
\noindent \textbf{EXAMPLE:}\\
\#\#\#Sentences:\\
1: The stock market experienced a sharp decline due to economic uncertainty.\\
2: Albert Einstein, a theoretical physicist, developed the theory of relativity.\\

\noindent \#\#\#Output:\\
\{"questions\_list": [["What sector faced a significant downturn because of economic uncertainty?", "Why did the stock market show a sudden decrease recently?", "What caused the sharp decline in the financial markets?"], ["Who is credited with developing the theory of relativity?", "What field was Albert Einstein associated with?", "What significant scientific theory did Albert Einstein develop?"]]\}\\

\noindent ==============\\
\#\#\#Sentences:\\
\{target ACUs\}\\
\#\#\#Output:\\

\noindent \textbf{\underline{Question Answering Prompt}}\\

\noindent \textbf{INSTRUCTION:} For each context-questions pairs, follow these steps:\\
1. Given the context, answer the following questions.\\
2. Consolidate all responses into a single concise sentence.\\

\noindent ==============\\
\noindent \textbf{EXAMPLE:}\\
Context 1: The stock market experienced a sharp decline due to economic uncertainty.\\
Q1: What sector faced a significant downturn because of economic uncertainty?\\
Q2: Why did the stock market show a sudden decrease recently?\\
Q3: What caused the sharp decline in the financial markets?\\
Context 2: Albert Einstein, a theoretical physicist, developed the theory of relativity.\\
Q1: Who is credited with developing the theory of relativity?\\
Q2: What field was Albert Einstein associated with?\\
Q3: What significant scientific theory did Albert Einstein develop?\\

\noindent \#\#\#Output:\\
\{"answers": ["The stock market experienced a sharp decline due to economic uncertainty.", " Albert Einstein, a theoretical physicist, developed the theory of relativity."]\}\\

\noindent ==============\\
\{context (similar ACUs) questions (generated questions) list\}\\
\#\#\#Output:\\

\subsection{Hyper-Parameter Selection}
\label{subsec:hyperparam_selection}
We describe the rationale of the hyperparameter selection in this section.

\subsubsection{Dynamic Salient Weight Adjustment}
\label{subsubsec:weight_appendix}
As introduced in Section \ref{subsec:acu_aggregation}, we adjust the weight of non-salient ACUs using a cubic function: $w_{ns} = \text{\textit{min}}(w_s, \alpha (p_s - \beta)^3 + \gamma)$, where $p_s$ represents the salience ratio of the document. This adjustment is designed to ensure that the overall {\nova} accurately reflects both the novelty and importance of the information within the document.

The parameter $\pmb{\alpha}$ controls the steepness of the cubic function, determining how sensitive the weight adjustment is to the salience ratio. A higher $\alpha$ results in a more pronounced adjustment, causing the weight of non-salient ACUs to decrease or increase more rapidly in response to very low or very high salience ratios. This sensitivity allows us to fine-tune how much emphasis is placed on non-salient ACUs depending on the distribution of salient information within the document.

The parameter $\pmb{\gamma}$ adjusts the midpoint on the $y$-axis, which corresponds to the general level of devaluation for non-salient ACUs, referred to as the "mean non-salience devaluation." For example, setting $\gamma = 0.7$ implies that, on average, non-salient ACUs are considered 70\% as important as salient ACUs, with further adjustments based on the document's salience ratio, as controlled by $\alpha$.

The parameter $\pmb{\beta}$ shifts the midpoint on the $x$-axis, determining the salience ratio at which the "mean non-salience devaluation" is applied. For instance, if $\beta = 0.5$ and $\gamma = 0.7$, then in documents where the salience ratio is less than 0.5, non-salient ACUs are assigned a lower weight than the mean devaluation of 0.7, with the rate of adjustment dictated by $\alpha$. Conversely, in documents with a salience ratio greater than 0.5, non-salient ACUs receive a higher weight than the mean devaluation, again with the rate of adjustment controlled by $\alpha$.

The choice of $\alpha$, $\beta$, and $\gamma$ depends on the specific dataset and application requirements. For the TAP-DLND 1.0 and APWSJ datasets used in our experiments, we performed a grid search with $\alpha \in [0, 2]$, $\beta \in [0, 0.8]$, and $\gamma \in [0.5, 1]$. The optimal hyperparameters for TAP-DLND 1.0 are found to be $\alpha = 0$, $\beta = 0.5$, and $\gamma = 1$, indicating no weight adjustment was necessary. For APWSJ, the optimal values are $\alpha = 1$, $\beta = 0.5$, and $\gamma = 0.7$. This discrepancy arises from the different standards and annotation approaches used in the two datasets. 

% For our human-annotated dataset, as discussed in Section \ref{subsec:gpt_performance}, the GPT-selected salient ACUs achieve a macro F1 score of 0.6 compared to human-annotated salient ACUs, which we consider to be suboptimal. The discrepancy likely arises from the different processes used to select salient ACUs: GPT first generates a summary and then selects the ACUs that appear in it, while human annotators must decide on the salience of each ACU in real-time. Due to this misalignment, we believe that applying weight adjustments based on GPT-selected salient ACUs could negatively impact the results when compared to {\nova}$_{\text{human}}$, which is computed using human-annotated novelty and salience. Consequently, we opt not to include the weight adjustment in this context.

The advantage of this weight adjustment scheme lies in its flexibility to control and incorporate both important and less important information when evaluating the overall novelty of a document. This provides {\nova} with an additional dimension, allowing it to assess not only the level of novelty but also the worthiness of the information within a target document.

\subsubsection{Similarity Thresholds}
\label{subsubsec:similarity_threshold}
We choose a threshold of 0.85 for embedding cosine similarity to determine whether two ACUs are almost identical because a higher threshold ensures that the two units are very close in semantic content. At this level, the embeddings are nearly overlapping, indicating that the ACUs convey virtually the same information with minimal variation. Conversely, a lower threshold of 0.6 is used to decide whether two ACUs are similar but not necessarily identical. This threshold allows for some semantic variation while still capturing a significant level of similarity, making it suitable for identifying ACUs that share related content or themes without being exact duplicates. These thresholds are selected based on empirical results, which demonstrate that they provide the best performance in distinguishing between near-duplicates and related content, thereby enabling a more nuanced analysis of a document's novelty and relevance.

\subsection{Correlation Statistics Interpretation}
\label{subsec:corr_interpretation}

\begin{table}[t]
\centering
\small
    \begin{tabular}{lccc}
    \toprule
    \textbf{Statistics $\rightarrow$} & \textbf{Pearson} & \textbf{Spearman} & \textbf{Kendall} \\
    \midrule
    \textbf{Strength $\downarrow$} \\
    Negligible & 0.00 & 0.00 & 0.00 \\
    Weak & 0.10 & 0.10 & 0.06 \\
    Moderate & 0.40 & 0.38 & 0.26 \\
    Strong & 0.70 & 0.68 & 0.49 \\
    Very Strong & 0.90 & 0.89 & 0.71 \\
    \bottomrule
    \end{tabular}
\caption{Cutoff values for the correlation statistics.}
\label{tab:corr_cutoffs}
\end{table}

Table \ref{tab:corr_cutoffs} details the cutoff values for the rank-based correlation statistics, which are based on the recommendations for the Pearson correlation by \citet{schober2018correlation}. Note that Point-Biserial statistics is a special case of Pearson correlation.

\subsection{Full {\nova} Correlation Results on Internal Data}
\label{subseec:corr_human_full}

\begin{table*}[ht]
\centering
\resizebox{0.95\textwidth}{!}
{
    \begin{tabular}{lccccccc}
    \toprule
    \textbf{Dataset $\rightarrow$} & \multicolumn{3}{c}{\textbf{w/ Salience}} && \multicolumn{3}{c}{\textbf{w/o Salience}} \\
    \textit{Score} $\rightarrow$ & \small{{\textbf{\nova}}$_{\text{\textbf{CosSim}}}$} & \small{{\textbf{\nova}}$_{\text{\textbf{NLI}}}$} & \small{{\textbf{\nova}}$_{\text{\textbf{QA}}}$} && \small{{\textbf{\nova}}$_{\text{\textbf{CosSim}}}$} & \small{{\textbf{\nova}}$_{\text{\textbf{NLI}}}$} & \small{{\textbf{\nova}}$_{\text{\textbf{QA}}}$} \\
    \cmidrule{2-4} \cmidrule{6-8}
    \textbf{Correlation $\downarrow$} &  &  &  &&  &  &  \\
    Pearson & 0.722$_{(3.1e-06)}$ & \textbf{0.835}$_{(2.9e-09)}$ & \underline{0.779}$_{(2.4e-07)}$ && 0.748$_{(8.6e-07)}$ & \textbf{0.920}$_{(9.6e-14)}$ & \underline{0.843}$_{(2.7e-09)}$ \\
    Spearman & \textbf{0.758}$_{(5.2e-07)}$ & \underline{0.567}$_{(7.3e-04)}$ & 0.562$_{(1.0e-04)}$ && \textbf{0.836}$_{(2.6e-09)}$ & 0.782$_{(1.2e-07)}$ & \underline{0.798}$_{(7.5e-08)}$ \\
    Kendall & \textbf{0.559}$_{(2.4e-05)}$ & \underline{0.423}$_{(1.4e-03)}$ & 0.409$_{(2.5e-03)}$ && \textbf{0.690}$_{(8.6e-07)}$ & \underline{0.687}$_{(1.9e-06)}$ & 0.643$_{(1.2e-05)}$ \\
    \bottomrule
    \end{tabular}
}
\caption{The correlations (statistics$_{(p\text{-value})}$) between automated {\nova} and {\nova}$_{\text{human}}$ computed from our internal annotated data, using different novelty evaluators.}
\label{tab:corr_human_full}
\end{table*}

Table \ref{tab:corr_human_full} details the full results of the correlations between fully automated {\nova} and {\nova}$_{\text{human}}$ on our annotated data, using different novelty evaluators.
\newpage
\section{Human Annotation}
\label{sec:human_details}

We provide the details of our human annotation process in this section.

\subsection{Annotation Instruction and Label Schema}
\label{subsec:annotation_instruction}
Following is the comprehensive annotation instruction and label schema we provide to the annotators.
\newline

\noindent \textbf{Instruction:} Articles are clustered and sorted by date within each cluster. Annotate the articles cluster by cluster, completing one cluster before moving on to the next. When annotating, reach each article in sequential order within its cluster. Memorize all information from the articles as you read. This is necessary for accurately judging the novelty of each ACU in subsequent articles within the same cluster. Novelty is only considered within the same cluster, not across different clusters.

First, read the news article carefully to understand the content and context of the entire article. Then label each ACU by the following steps. 
\newline

\noindent \textbf{Step 1: Assessing Correctness and Redundancy}

\noindent Evaluate each ACU within the context of the article to determine the correctness. Determine the redundancy of the ACU by comparing it with the previous ACUs within the same article.
\newline

\noindent \textbf{Label Schema}

\noindent \textbf{\textit{Correctness}}

\noindent \underline{correct:} The ACU is accurate and logically consistent within the context of the article.

\noindent \underline{incorrect:} The ACU contains incorrect information, errors, illogical, or LLM hallucinations.
\newline

\noindent \textbf{\textit{Redundancy}}

\noindent \underline{redundant:} The ACU

(a) is a direct repeat or rephrase of a previous ACU within the current article.

(b) does not convey any meaningful information. This is usually the case where the ACU describes the metadata of the article. For instance, an ACU such as "The article is written by xxx" or "The publish date of the article is xxx" should be marked as redundant.

\noindent \underline{not-redundant:} The ACU provides new unique and meaningful information within the current article. If an ACU is partially new, it is also considered not-redundant.
\newline

\noindent \textbf{Step 2: Assessing Novelty and Salience} (Only for Correct and Not Redundant ACUs)

\noindent Evaluate the novelty of each ACU by comparing it with the previous articles in the same clusters to check if all information in the ACU is already known. For each ACU, you will be shown the top 5 similar ACUs from previous articles. For the first article within the cluster, no similar ACUs will be shown as we assume there are no older articles to compare with. Therefore, all correct and not redundant ACUs in the first article should be considered novel.
Use similar ACUs only as a reference.

\noindent \underline{Situation 1:} Information not in the top 5 similar ACUs does not necessarily mean that it is not mentioned in previous articles. Try your best to memorize what you've read and always go back to the original article to verify if you recall something you've read but is not in the top 5 similar ACUs.

\noindent \underline{Situation 2:} If an article is different in topic/domain than the previous articles, the top 5 similar ACUs might not be useful at all. Please always refer back to the original articles to check for detailed information.
Assess whether the information in the ACU is crucial for understanding the main points of the article to determine the salience of the ACU.
\newline

\noindent \textbf{Label Schema}

\noindent \textbf{\textit{Novelty}}

\noindent \underline{novel:} The ACU introduces some new information that is not present in previous article. If an ACU is partially new, it is also considered novel.

\noindent \underline{not-novel:} The ACU does not introduce any new information in the sense that all information mentioned in this ACU has been mentioned in older articles within the same cluster. Only consider inter-article novelty, not intra-article novelty -- an ACU should only be annotated as "not-novel" if all information has been mentioned in previous articles within the same cluster. If an ACU introduces the exact same information as an earlier ACU within the same article, it should be labeled as "redundant".
\newline

\noindent \textbf{\textit{Salience}}

\noindent \underline{salient:} The ACU contains the essential information that you would include in a summary of the article – label the ACU as salient if you think it is an essential information to convey the main point of the article.

\noindent \underline{non-salient:} If the ACU does not contain essential information for the summary.

\subsection{Annotation Quality}
\label{subsec:annotation_quality}

\begin{table}[ht]
\centering
\resizebox{0.48\textwidth}{!}
{
    \begin{tabular}{lcccc}
    \toprule
    \textbf{Metric $\rightarrow$} & \textbf{Precision} & \textbf{Recall} & \textbf{F1-Score} & \textbf{Support} \\
    \cmidrule{2-5}
    \textbf{Class $\downarrow$} \\
    Non-Novel & 0.00 & 0.00 & 0.00 & 0 \\
    Novel & 1.00 & 0.99 & 1.00 & 222 \\
    \midrule
    Accuracy & & & 0.99 & 222 \\
    Weighted Avg & 1.00 & 0.99 & 1.00 & 222 \\
    \bottomrule
    \end{tabular}
}
\caption{The classification report of human annotation on \textbf{expected novel} ACUs.}
\label{tab:qr_result_novel}
\end{table}

\begin{table}[ht]
\centering
\resizebox{0.48\textwidth}{!}
{
    \begin{tabular}{lcccc}
    \toprule
    \textbf{Metric $\rightarrow$} & \textbf{Precision} & \textbf{Recall} & \textbf{F1-Score} & \textbf{Support} \\
    \cmidrule{2-5}
    \textbf{Class $\downarrow$} \\
    Non-Novel & 1.00 & 0.82 & 0.90 & 22 \\
    Novel & 0.00 & 0.00 & 0.00 & 0 \\
    \midrule
    Accuracy & & & 0.82 & 22 \\
    Weighted Avg & 1.00 & 0.82 & 0.90 & 22 \\
    \bottomrule
    \end{tabular}
}
\caption{The classification report of human annotation on \textbf{expected non-novel} ACUs.}
\label{tab:qr_result_nonnovel}
\end{table}

We have two annotators independently perform the entire annotation task. After completing their annotations, they meet to discuss and resolve any conflicting labels, ensuring consensus on the final results. To further ensure the quality of the annotations, we discreetly create and insert three synthetic articles as quality control samples without informing the annotators. Two of these articles are complete paraphrases of previous articles within a cluster and are added to the end of the cluster; for these, all ACUs are expected to be non-novel. The third article is manually written as a completely new piece, unrelated to any other articles in the cluster, where all ACUs are expected to be novel. Additionally, for the first article in each cluster, all ACUs are also expected to be novel. As shown in Tables \ref{tab:qr_result_novel} and \ref{tab:qr_result_nonnovel}, the human annotation achieves a weighted F1 score of 1.0 on the expected novel ACUs and 0.9 on the expected non-novel ACUs, indicating the high quality of the annotation process.

\end{document}